\documentclass[letterpaper]{article} % DO NOT CHANGE THIS
\usepackage{aaai25}  % DO NOT CHANGE THIS
\usepackage{times}  % DO NOT CHANGE THIS
\usepackage{helvet}  % DO NOT CHANGE THIS
\usepackage{courier}  % DO NOT CHANGE THIS
\usepackage[hyphens]{url}  % DO NOT CHANGE THIS
\usepackage{graphicx} % DO NOT CHANGE THIS
\usepackage{natbib}  % DO NOT CHANGE THIS AND DO NOT ADD ANY OPTIONS TO IT
\usepackage{caption} % DO NOT CHANGE THIS AND DO NOT ADD ANY OPTIONS TO IT
\usepackage{algorithm}
\usepackage{algorithmic}

\usepackage{newfloat}
\usepackage{listings}
\DeclareCaptionStyle{ruled}{labelfont=normalfont,labelsep=colon,strut=off} % DO NOT CHANGE THIS
\floatstyle{ruled}
\newfloat{listing}{tb}{lst}{}
\floatname{listing}{Listing}
\usepackage{adjustbox}
\usepackage{tcolorbox}
\usepackage{xcolor}
\usepackage{colortbl}
\usepackage{booktabs}
\usepackage{multirow}
\usepackage{makecell}
\usepackage{subcaption}
\usepackage{amsmath,amssymb,amsfonts}
\usepackage{bm}
\usepackage{placeins}

\newcommand{\ie}{\textit{i}.\textit{e}.}
\newcommand{\eg}{\textit{e}.\textit{g}.}

\newcommand{\z}{\mathbf{z}}
\newcommand{\cond}{c}

\newcommand{\zk}{z_{t}^{\state+t}}
\newcommand{\zkm}{z_{t-1}^{\state+t}}
\newcommand{\ttoT}[1]{\left\{#1\right\}_{t=1}^{T}}
\newcommand{\itof}[1]{\left\{#1\right\}_{i=1}^{f}}
\newcommand{\state}{\tau}

\newcommand{\zsttm}{z^{\state+t}_{t-1}}

\title{Ouroboros-Diffusion: Exploring Consistent Content Generation \\
	in Tuning-free Long Video Diffusion\thanks{This work was performed at HiDream.ai.}}

\author{
    Jingyuan Chen\textsuperscript{\rm 1},
    Fuchen Long\textsuperscript{\rm 2},
    Jie An\textsuperscript{\rm 1},
    Zhaofan Qiu\textsuperscript{\rm 2},
    Ting Yao\textsuperscript{\rm 2},
    Jiebo Luo\textsuperscript{\rm 1},
    Tao Mei\textsuperscript{\rm 2}
}
\affiliations{
    \textsuperscript{\rm 1} University of Rochester, Rochester, NY USA \\
    \textsuperscript{\rm 2} HiDream.ai Inc. \\
	{jchen157@u.rochester.edu},~~{longfuchen@hidream.ai},~~{jan6@cs.rochester.edu} \\
	{\{qiuzhaofan, tiyao\}@hidream.ai},~~{jluo@cs.rochester.edu},~~{tmei@hidream.ai} \\
}

\usepackage{bibentry}
\begin{document}

\maketitle

\begin{abstract}
The first-in-first-out (FIFO) video diffusion, built on a pre-trained text-to-video model, has recently emerged as an effective approach for tuning-free long video generation.
This technique maintains a queue of video frames with progressively increasing noise, continuously producing clean frames at the queue’s head while Gaussian noise is enqueued at the tail. 
However, FIFO-Diffusion often struggles to keep long-range temporal consistency in the generated videos due to the lack of correspondence modeling across frames.
In this paper, we propose Ouroboros-Diffusion, a novel video denoising framework designed to enhance structural and content (subject) consistency, enabling the generation of consistent videos of arbitrary length.
Specifically, we introduce a new latent sampling technique at the queue tail to improve structural consistency, ensuring perceptually smooth transitions among frames.
To enhance subject consistency, we devise a Subject-Aware Cross-Frame Attention (SACFA) mechanism, which aligns subjects across frames within short segments to achieve better visual coherence. Furthermore, we introduce self-recurrent guidance. This technique leverages information from all previous cleaner frames at the front of the queue to guide the denoising of noisier frames at the end, fostering rich and contextual global information interaction.
Extensive experiments of long video generation on the VBench benchmark demonstrate the superiority of our Ouroboros-Diffusion, particularly in terms of subject consistency, motion smoothness, and temporal consistency.
\end{abstract}


\section{Introduction}
With the rise of artificial visual content generation technologies, significant breakthroughs have been made in video diffusion~\cite{blattmann2023videoldm,peng2024aaai,ma2024aaai,long2024eccv}. 
However, most current video diffusion models are trained on short clips (\eg, 16 frames), a significant challenge when scaling to long video generation. Instead of relying on extensive training data to extend video length, our work focuses on leveraging a pre-trained video diffusion model to generate long videos without extensive training data or fine-tuning. Additionally, we place a strong emphasis on content consistency to enhance both the visual and motion quality of long videos in diffusion.

\begin{figure}[t!]
	\centering
	\includegraphics[width=0.99\linewidth]{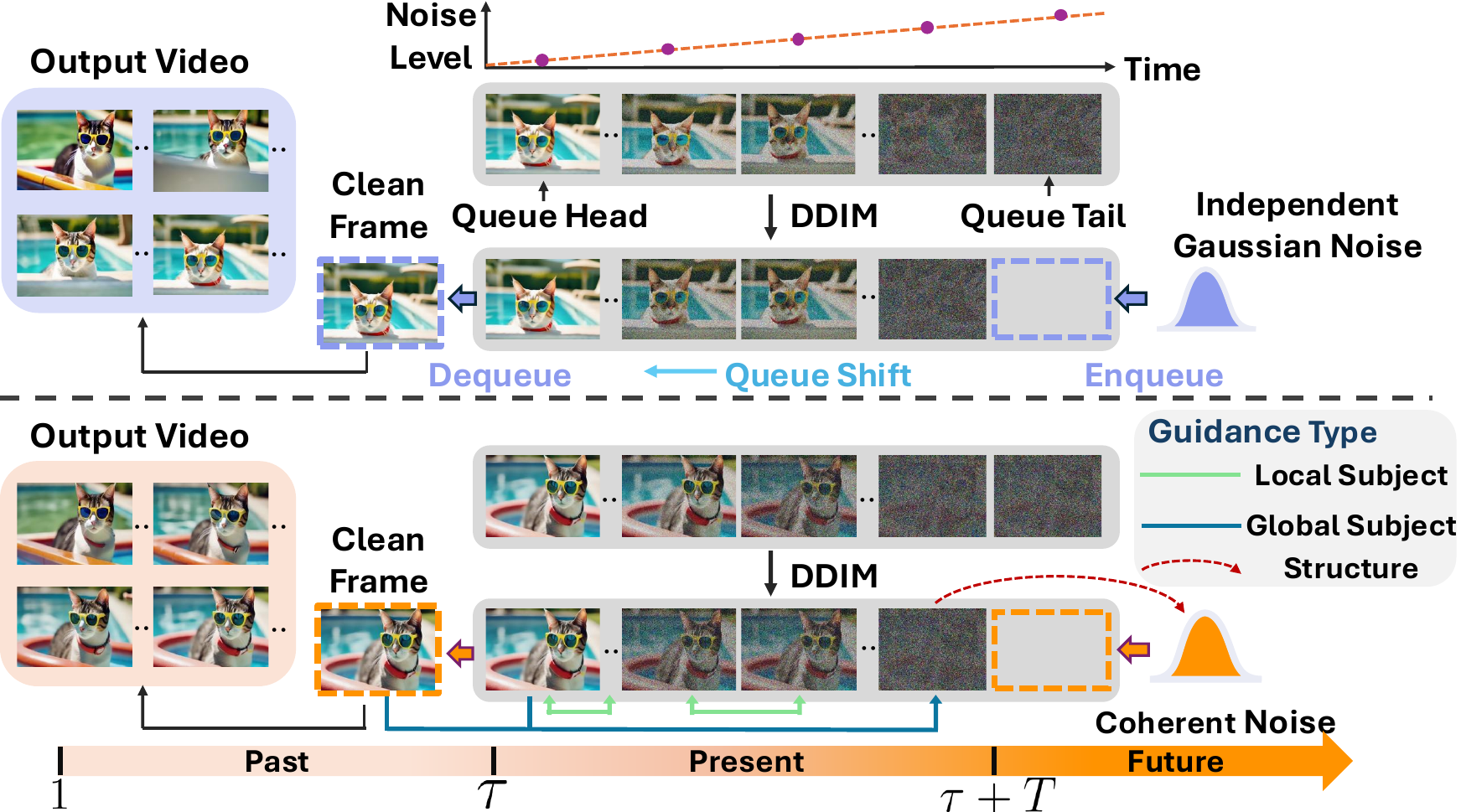}
	\caption{Illustration of FIFO-Diffusion~\cite{kim2024fifo} (top) and our Ouroboros-Diffusion (bottom) for tuning-free long video generation.}
	\label{fig:intro_fig}
\end{figure}

The recent first-in-first-out diffusion~\cite{kim2024fifo} has successfully employed the pre-trained video diffusion model for infinite frame generation.  Compared
to general video diffusion that all frames share the same
noise level, the diagonal denoising approach proposed by \citet{kim2024fifo} maintains a queue of frames with progressively increasing noise levels, enabling frame-by-frame generation at each step.
The upper part of Figure \ref{fig:intro_fig} conceptualizes the diagonal denoising process.
In this process, a fully denoised frame at the queue head is popped out, while a random noisy latent is pushed to the tail. The latent enqueue-dequeue cycle allows for the incremental generation of video frames.
However, the independently enqueued Gaussian noise can lead to content discrepancies between the frame latents near the tail of the queue.
Moreover, FIFO-Diffusion fails to utilize visual information contained in the earlier denoised frames, which exacerbates temporal inconsistency in continuous video diffusion.
For instance, the appearance of the cat changes significantly in the output video of Figure \ref{fig:intro_fig}.
Our work addresses this issue with FIFO-Diffusion from two key perspectives: structural and subject consistency. To this end, we introduce a novel denoising framework termed Ouroboros-Diffusion, designed for tuning-free long video generation. 
Inspired by the ancient symbol of Ouroboros—a serpent or dragon eating its own tail, symbolizing wholeness and self-renewal—our framework embodies these concepts by seamlessly integrating information across time.
The design of Ouroboros-Diffusion is guided by three core principles targeting distinct information flows to improve structural and subject consistency: \textit{present infers future, present influences present, and past informs present}, as depicted in Figure \ref{fig:intro_fig}.
To enhance structural consistency, we address the critical step of enqueueing new tail noise, which can lead to structural incoherence. From a video continuation perspective, this step involves sampling a \textit{future} frame that should maintain visual structure continuity with previous frames. 
To achieve this, we propose inferring the \textit{future} frame from the \textit{present} frame in the denoising queue by exploiting the low-frequency component connection between them. 
Specifically, instead of initializing the tail latent with Gaussian noise, Ouroboros-Diffusion extracts the low-frequency component from the second-to-last frame latent using Fast Fourier Transform (FFT) and combines it with the high-frequency part of random noise to create the enqueued latent. The low-frequency component preserves layout information for overall video consistency, while the high-frequency component introduces necessary video dynamics.
For subject consistency, we consider the semantic dependencies of long videos from two angles: within the current denoising queue and between previously generated clear frames and the current queue. 
To enhance subject temporal coherence within the \textit{present} queue, we extend self-attention across frames through a Subject-Aware Cross-Frame Attention (SACFA) module. This module leverages segmented subject regions from the cross-attention map to extract subject tokens in each frame, which serve as auxiliary contexts for subject alignment. These tokens are then stored in a subject feature bank.
To model longer-range subject dependencies, we introduce a long-term memory of \textit{past} subjects to guide the appearance of the \textit{present} subject. Specifically, Ouroboros-Diffusion utilizes the long-term memory derived from the frame at the head of the queue to guide the denoising of noisier frames near the tail, optimizing the latent through a subject-aware gradient during video denoising.

The main contribution of this work is the proposal of Ouroboros-Diffusion to address content consistency in turning-free long video generation.
Our solution elegantly explored how diagonal denoising could benefit from low-frequency content preservation, and how to preserve subject consistency through cross-frame attention and gradient-based latent optimization in diffusion.  
Extensive experiments on VBench verify the effectiveness of our proposal in terms of both visual and motion quality.

\section{Related Work}
\paragraph{Text-to-Video Diffusion Models.} 
The great success of text-to-video (T2V) diffusion models~\cite{ho2022imagen,voleti2022mcvd,villegas2022phenaki,2023videocomposer,yin2023dragnuwa,chen2023videocrafter1, zhang2024trip,guo2024animatediff} for video generation based on text prompts has been witnessed in recent years.
VDM~\cite{ho2022vdm} is one of the early works that combines spatial and temporal attention to construct space-time factorized UNet for video synthesis.
Later in Make-A-Video~\cite{singer2022make}, the prior knowledge of text-to-image diffusion models is explored in video diffusion and the 2D-UNet is extended with the temporal modules (\eg, temporal convolution and self-attention~\cite{long2022sifa}) for motion modeling~\cite{long2019gaussian,long2022dynamic,long2023BCN}.
The advances~\cite{an2023latent, blattmann2023videoldm, 2023tune, chen2024sateco} further execute the video synthesis on latent space and push the boundaries of high-resolution video generation.
Inspired by the impressive performances of Diffusion Transformers (DiTs)~\cite{DiT,SiT}, the spatial-temporal transformer architecture starts to emerge in video diffusion~\cite{2023cogvideo,easyanimate}. 
Here, we choose VideoCrafter2~\cite{chen2024videocrafter2} as a backbone text-to-video model for long video generation.

\paragraph{Long Video Diffusion.}
Despite the achievements of text-to-video diffusion, long video generation is still a grand challenge.
Existing works have explored two strategies, \ie, tuning-based and tuning-free long video diffusion.
Typically, the tuning-based methods~\cite{yin2023nuwa, henschel2024streamingt2v,tian2024videotetris,videolavit} usually exploits an auto-regressive manner which leverages the information of past generated frames to guide the synthesis of current frames.
Nevertheless, tuning the auto-regressive video diffusion model usually involves a huge computational cost.
To overcome this limitation, tuning-free approaches~\cite{wang2023gen,qiu2023freenoise,kim2024fifo,oh2023mtvg,tan2024video} try to adopt a pre-trained basic video diffusion model with the power of short-clip synthesis to generate multiple frames with temporal coherence.
\citet{qiu2023freenoise} introduces a sliding-window temporal attention mechanism to simultaneously denoising all frames for keeping temporal consistency.
Recent advance FIFO-Diffusion~\cite{kim2024fifo} proposes to store frames with different noise levels in a queue, and continuously output one clean frame from the head and enqueue Gaussian noise at the tail in each denoising step.
However, the model still faces the temporal flickering issue due to insufficient long-range modeling and discrepancy arising from enqueueing Gaussian tail noise.

\paragraph{Guidance in Denoising.}
In the procedure of image/video denoising, various guidance (\eg, visual tokens in attention \cite{tewel2024training,Jain2024CVPR} and gradient guidance ~\cite{Meng2022SD,Epstein2023self,an2023openleaf,chen2024layout}) has been investigated to control the content generation.
For instance, ConsiStory~\cite{tewel2024training} takes the visual tokens from the reference image to facilitate the content alignment across images in a batch, while FreeDoM~\cite{yu2023freedom} optimizes the denoising procedure with the gradient of the target energy function.

\paragraph{}
In short, our work exploits a basic T2V model for long video generation without tuning.
The proposed Ouroboros-Diffusion contributes by not only studying how the low-frequency component in latent influences the structural consistency, but also how the guidance in video denoising can be better leveraged to achieve frame-level subject consistency.

\begin{figure*}[t!]
	\centering
	\includegraphics[width=0.99\linewidth]{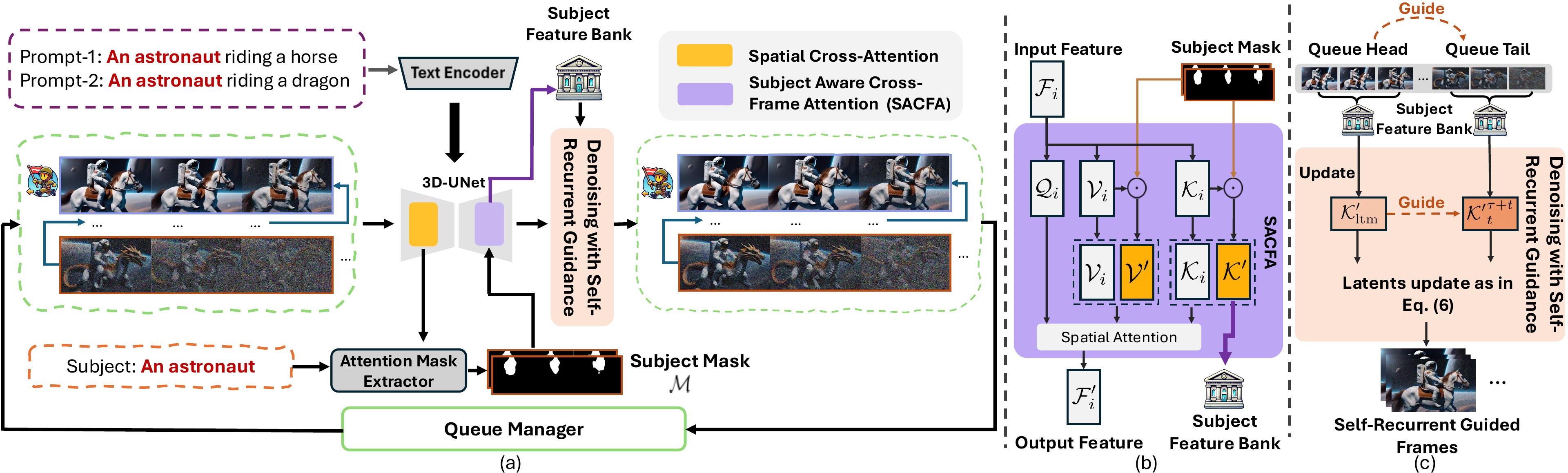}
	\caption{
		An overview of our Ouroboros-Diffusion. The whole framework (a) contains three key components: coherent tail latent sampling in queue manager , (b) Subject-Aware Cross-frame Attention (SACFA), and (c) self-recurrent guidance. 
		The coherent tail latent sampling in queue manager derives the enqueued frame latents at the queue tail to improve structural consistency.
		The Subject-Aware Cross-frame Attention (SACFA) aligns subjects across frames within short segments for better visual coherence. 
		The self-recurrent guidance leverages information from all historical cleaner frames to guide the denoising of noisier frames, fostering rich and contextual global information interaction.}
	\label{fig:framework}
\end{figure*}

\section{Preliminaries: Video Denoising Approach}

\paragraph{Parallel Denoising.} 
Latent Video Diffusion Models (LVDMs) perform diffusion processes in the latent space for efficient video generation.
Given a video latent \(\z_{t} \in \mathbb{R}^{f \times c \times h \times w}\) at timestep \(t \in \left[1,\ldots, T\right]\), where \(\z_{t}\) consists of \(f\) frame latents: \(\z_{t} = \itof{z_t^i}\), \(T\) represents the total number of denoising steps used in the sampling process. 
Conventionally, LVDMs adopt a parallel denoising approach, where \(\z_t\) is iteratively denoised to obtain the clean video latent \(\z_0\). In parallel denoising, the noise level remains consistent across all frames throughout the denoising process.
The parallel denoising is formulated as:
\begin{equation}
	\z_{t-1} = \boldsymbol{\Psi} \left(\z_t, t, \bm{\epsilon}_{\theta}(\z_t,t,\cond)\right)~,
	\label{eq:denoise_sync}
\end{equation}
where $\boldsymbol{\Psi}(\cdot)$ denotes the sampler such as DDIM and \(\bm{\epsilon}_{\theta}\) denotes a spatial-temporal UNet that predicts the added noise at each denoising step, conditioned on the text embedding \(\cond\).

\paragraph{Diagonal Denoising.}
Unlike the conventional parallel denoising, FIFO-Diffusion~\cite{kim2024fifo} introduces a diagonal denoising technique that sequentially produces clear frame latents.
This is achieved by employing a fixed-length queue of frame latents with progressively increasing noise levels (\ie, 1 to \(T\)), as illustrated in Figure \ref{fig:intro_fig}.
In the time step \(\state\) of diagonal denoising, $\state-1$ frames have already been generated.
We denote \(\mathbf{Q}^\state = \ttoT{z_t^{\state+t}}\) as all frame latents in the queue. 
Here, \(z_t^{\state+t}\) is the (\(\state+t\))-th frame latent with noise level $t$.
The denoising step is reformulated as:
\begin{equation}
	\ttoT{\zsttm} =\boldsymbol\Psi(\mathbf{Q}^\state, \ttoT{t}, \bm{\epsilon}_{\theta}(\mathbf{Q}^\state,
	\ttoT{t}, \cond))~,
\end{equation}
\noindent
where \(\ttoT{\zsttm}\) denotes the one-step denoised \(\mathbf{Q}^\tau\).
After the denoising step, the first frame latent $z_0^{\state}$ in the queue becomes a clear latent and is dequeued from the head.
A newly sampled Gaussian noise is then enqueued at the tail, making the queue $\mathbf{Q}^{\state}$ to transition to $\mathbf{Q}^{\state+1}$. Iteratively performing this enqueue-dequeue process allows for video generation in a frame-by-frame manner.
When the queue size \(T\) exceeds the frame capacity \(f\) of the base video diffusion model, the frame latents are denoised window-by-window with the basic temporal length \(f\).

\section{Methodology}
Figure \ref{fig:framework} provides an overview of the Ouroboros-Diffusion framework.
To address the limitations of diagonal denoising, we model three distinct types of information flow to achieve structural and subject consistency.
Coherent tail latent sampling ensures smooth transitions by using the second-to-last latent as the guidance, enabling \textit{present} to infer \textit{future}. 
SACFA enhances subject consistency by extending spatial self-attention with subject contexts from neighboring frames, allowing mutual influence between \textit{present} frames. 
Self-recurrent guidance further elevates long-range subject coherence by leveraging past subject memory derived from the head of the queue to inform the denoising of the tail, demonstrating how the \textit{past} informs the \textit{present}.
In this section, we first discuss the limitations of the FIFO-Diffusion as the motivation of the proposed method. Then we introduce details of our Ouroboros-Diffusion.

\subsection{Limitation of Diagonal Denoising}
The FIFO-Diffusion~\cite{kim2024fifo} enables the generation of videos with infinite frames. However, the subject consistency of the videos produced by diagonal denoising is often compromised due to the following two limitations:

\paragraph{Lack of Global Consistency Modeling.}
In the design of diagonal denoising, the global visual consistency is not explicitly considered during the diffusion process.
Once frames are fully denoised, they are dequeued and no longer contribute to the generation of subsequent frames. This leads to the underutilization of embedded subject information(\eg, semantics, motion). Consequently, later frames cannot reference the information from previously generated frames, which causes the visual appearance of backgrounds and objects to gradually shift during continuous frame denoising, resulting in inconsistencies over time.

\begin{figure}[ht!]
	\centering
	\includegraphics[width=0.99\linewidth]{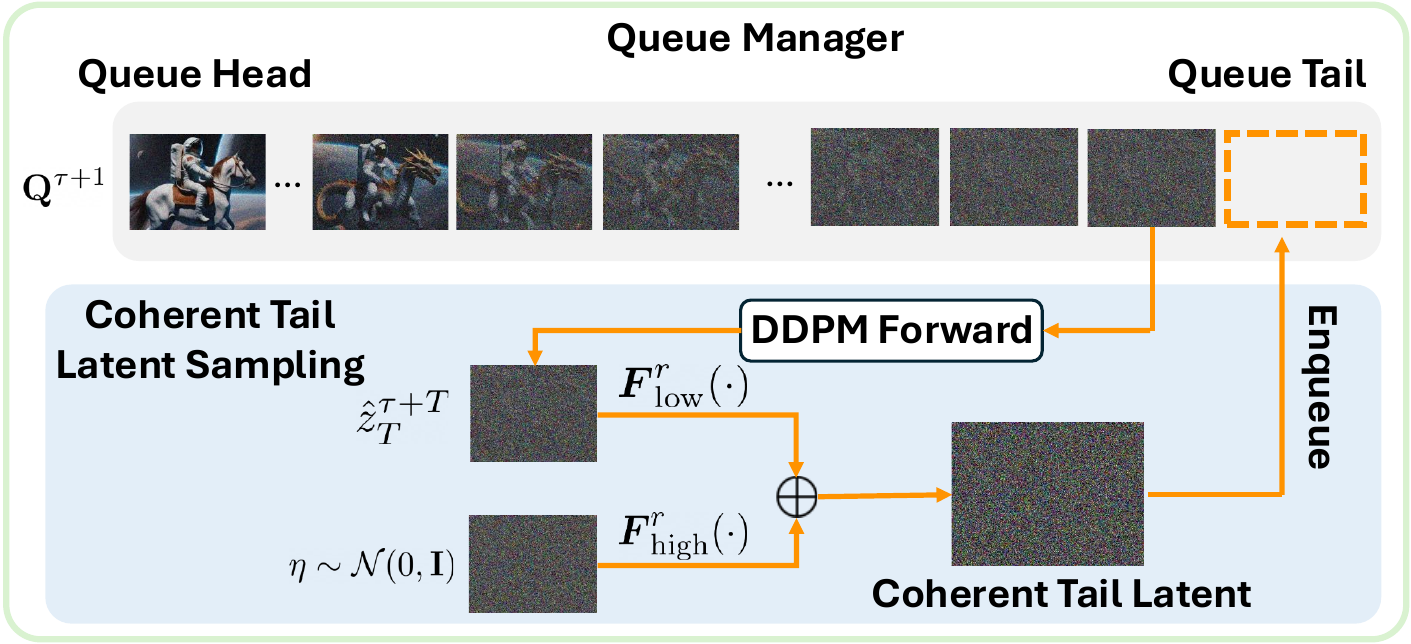}
	\caption{The detailed illustration of coherent tail latent sampling in the queue manager.}
	\label{fig:noise_manager}
\end{figure}

\paragraph{Latent Discrepancy Near the Tail.}
Another limitation arises in the enqueue-dequeue denoising process, specifically in the construction of the tail latent. The tail latent is typically sampled from a standard Gaussian distribution, which contains pure noise without any visual information. Meanwhile, neighboring latents have already undergone partial denoising in previous steps, introducing some degree of visual content. This creates a discrepancy between the visual information in the tail latent and its neighboring latents, leading to inconsistencies. As a result, the model may struggle to reconcile these differences, often resulting in frame flickering during video generation.

\paragraph{}
These limitations in diagonal denoising motivate us to devise targeted strategies for improvement.

\subsection{Coherent Tail Latent Sampling}
After completing a DDIM sampling step for queue \(\mathbf{Q}^\state\), a clean latent is removed from the queue head, leaving a vacant spot at the tail with a noise level of $T$. 
Instead of sampling Gaussian noise as in \citet{kim2024fifo}, we propose coherent tail latent sampling to retain similar structural information by using the second-to-last latent \(z_{T-1}^{\state+T}\) as a structural guidance.
To ensure the structure similarity between the last two latents, a straightforward approach is to directly apply noise to the second-to-last frame latent and use it as the new tail latent.
However, we discover that this approach results in generated videos with limited dynamics due to the excessive similarities in visual content.
Recent advances \cite{wu2023freeinit, everaert2024exploiting} indicate that the low-frequency component in the latent space primarily corresponds to the layout and overall structure in the pixel space. As shown in Figure \ref{fig:noise_manager}, we apply a 2D low-pass filter to extract the low-frequency component of the re-noised latent \(\hat{z}_{T}^{\state+T}\), preserving the layout as the base latent, and introduce dynamics by adding the high-frequency component of a randomly sampled Gaussian noise~$\eta$.
The coherent tail latent sampling is formulated as:
\begin{equation}
	z_{T}^{\state+1+T} = \boldsymbol{F}_\text{low}^{r}(\hat{z}_{T}^{\state+T})+\boldsymbol{F}_\text{high}^{r}(\eta)~,
\end{equation}
where $\boldsymbol{F}_\text{low}^{r}(\cdot)$ and $\boldsymbol{F}_\text{high}^{r}(\cdot)$ denotes the low-pass and high-pass filter functions with a threshold $r$, respectively. 
The coherent tail latent allows for consistent yet dynamic visual continuation, ensuring a smooth transition from \(\mathbf{Q}^{\state}\) to \(\mathbf{Q}^{\state+1}\).
In this way, the \textit{future} is faithfully inferred from the \textit{present} information within the queue.

\subsection{Subject-Aware Cross-Frame Attention}
To improve subject consistency during denoising, we propose Subject-Aware Cross-Frame Attention (SACFA), which extends the vanilla spatial attention layer to incorporate subject tokens from multiple frames, enhancing visual alignment across frames with enriched subject context.

\paragraph{Subject Mask Construction. }
Central to SACFA is a subject token masking mechanism, which relies on segmenting the subject within each frame. To obtain the segmentation, key subject words are first extracted from a prompt using GPT-4o \cite{openai2023gpt4}. These words are then tokenized and encoded into the embedding \(\mathcal{C}_\text{subj}\) using the CLIP \cite{radford2021learning}  text encoder.
The subject tokens are fed into the linear projection layer of each cross-attention layer to form the text subject key \(\mathcal{K}_\text{subj}\). Subject-related attention maps are obtained by computing the attention between the query \(\mathcal{Q}\) and \(\mathcal{K}_\text{subj}\) for each cross-attention layer. These maps are averaged across the token dimension and converted into a binary subject mask \(\mathcal{M}\) using Otsu’s method. Finally, subject masks of different resolutions are interpolated to a uniform resolution and averaged, resulting in the final subject mask.

\paragraph{Attention Processing of SACFA.}
To enhance temporal subject consistency across neighboring frames, we extend the spatial self-attention layer into a subject-aware cross-frame approach. The essence is to enable each frame to incorporate subject visual content from other frames when modeling its own appearance. To achieve this, we start by applying subject-specific masks to the keys and values of all frames to extract relevant visual content. These masked keys and values are then concatenated across frames to form subject-aware cross-frame keys and values, denoted as \(\mathcal{K}^\prime\) and \(\mathcal{V}^\prime\), with the shape \(fh^\prime w^\prime \times d\). This process creates a collection of subject references that capture the subject information across multiple frames. Finally, \(\mathcal{K}^\prime\) and \(\mathcal{V}^\prime\) are concatenated with the regular key \(\mathcal{K}_i\) and value \(\mathcal{V}_i\) of the target frame \(i\). 
The attention processing in SACFA for the \(i\)-th frame is then computed as:
\begin{equation}
	\mathcal{F}'_i = \text{Softmax} \left( \frac{\mathcal{Q}_i \cdot \begin{bmatrix} \mathcal{K}_i, \mathcal{K^\prime} \end{bmatrix}^{\top}}{\sqrt{d}} \right) 
	\cdot \begin{bmatrix} \mathcal{V}_i, \mathcal{V^\prime} \end{bmatrix}.
	\label{eq:SACFA}
\end{equation}
SACFA strengthens local subject correspondence by allowing subject information from neighboring frames to influence current frames, ensuring consistent subject representation. The subject-related key \(\mathcal{K^\prime}\) is then stored in a subject feature bank, forming a subject memory that serves as the foundation for subsequent self-recurrent guidance.

\begin{figure*}[t!]
	\centering
	\includegraphics[width=0.90\linewidth]{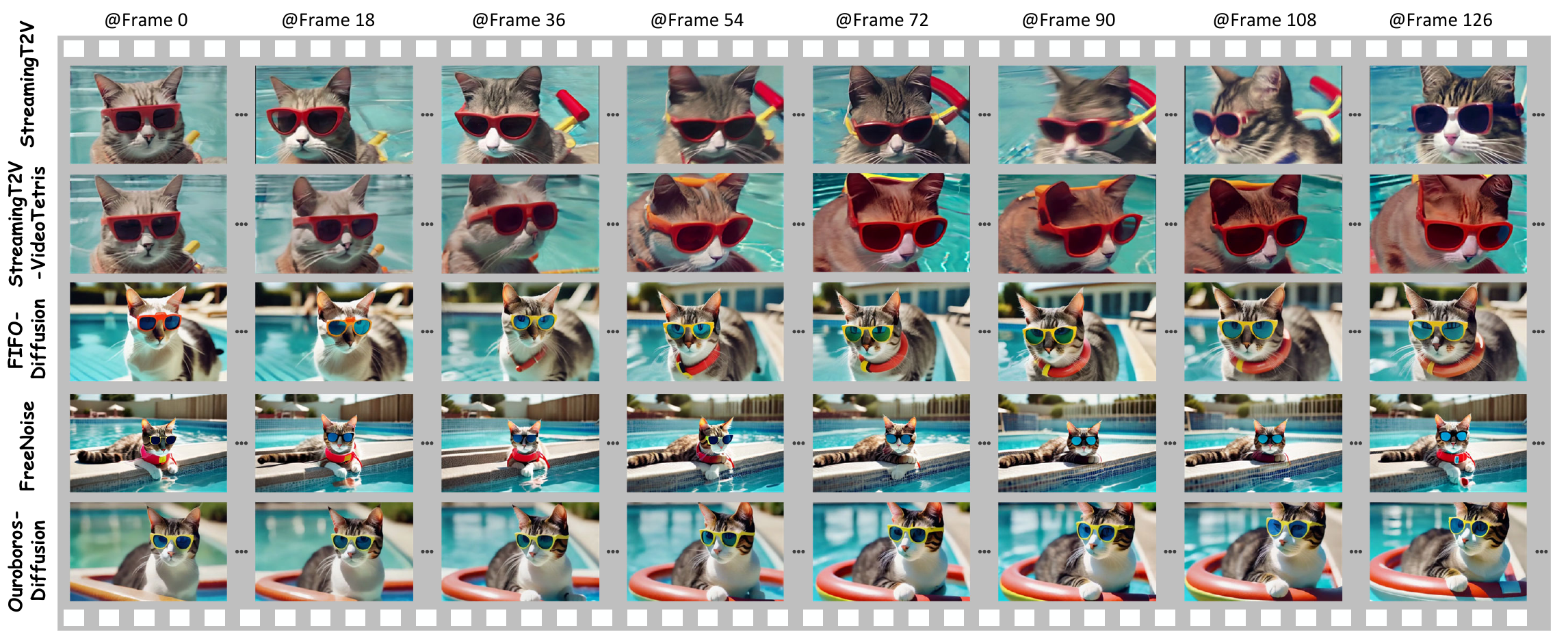}
	\caption{Visual examples of single-scene long video generation by different approaches. The text prompt is ``A cat wearing sunglasses and working as a lifeguard at a pool.''}
	\label{fig:single_case}
\end{figure*}

\begin{table*}[t]
	\centering
	\setlength{\tabcolsep}{0.7cm}{
		\resizebox{\linewidth}{!}{
			\begin{tabular}{lcccccc}
				\toprule
				\textbf{Approach} & 
				\makecell[c]{\textbf{Subject}\\ \textbf{Consistency}}$\boldsymbol\uparrow$ & 
				\makecell[c]{\textbf{Background}\\ \textbf{Consistency}}$\boldsymbol\uparrow$ & 
				\makecell[c]{\textbf{Motion}\\ \textbf{Smoothness}}$\boldsymbol\uparrow$ & 
				\makecell[c]{\textbf{Temporal}\\ \textbf{Flickering}}$\boldsymbol\uparrow$ & 
				\makecell[c]{\textbf{Aesthetic}\\ \textbf{Quality}}$\boldsymbol\uparrow$ & \\
				\midrule
				\multicolumn{1}{l|}{StreamingT2V~\cite{henschel2024streamingt2v}}   & 90.70     & 95.46     & 97.34     & 95.93     & 54.98  \\
				\multicolumn{1}{l|}{StreamingT2V-VideoTetris~\cite{tian2024videotetris}}        
				&  89.06    & 94.80     & 96.79     & 95.30     & 52.89  \\
				\multicolumn{1}{l|}{FIFO-Diffusion~\cite{kim2024fifo}}                        & 94.04    & 96.08    & 95.88  & 93.38  & 59.06 \\
				\multicolumn{1}{l|}{FreeNoise~\cite{qiu2023freenoise}}              & 94.50    & 96.45    & 95.42  & 93.62   & 59.32    \\ \midrule
				\multicolumn{1}{l|}{Ouroboros-Diffusion}                                      & \textbf{96.06}   & \textbf{96.90}    &  \textbf{97.73}  &  \textbf{96.12} &  \textbf{59.89} \\
				\bottomrule
			\end{tabular}
	}}
	\caption{Single-scene video generation performances on VBench. For each video, 128 frames are synthesized for evaluation.} 
	\label{tab:sig_text_exp}
\end{table*}

\subsection{Self-Recurrent Guidance}
We introduce self-recurrent guidance to leverage the \textit{past} subject features to guide the \textit{present} denoising steps.
Central to this approach is the Subject Feature Bank, which stores the long-term memory of video subjects.
The Subject Feature Bank is initialized using the averaged subject-masked keys \(\mathcal{K^\prime}\) from the first \(f\) cleaner frame latents of the video sequence, denoted as \(\mathcal{K}^\prime_\text{ltm}\). These initial $f$ frames are particularly valuable as they contain clearer and more critical visual information, making them an essential basis for the construction of long-term memory.

After each denoising step of the queue, the bank is updated using an exponential moving average as follows:
\begin{equation}
	\mathcal{K}^\prime_\text{ltm} \leftarrow \lambda \cdot \mathcal{K}^\prime_\text{ltm} + \frac{1-\lambda}{f} \cdot \sum_{t=1}^f{\mathcal{K^\prime}_{t}^{\state+t}}~,
\end{equation}
where $\lambda$ denotes the strength of memorization and only the first \(f\) frames in the queue contribute to the update.

Then, we exploit the subject tokens from the feature bank as a reference to minimize the gradient of the subject discrepancy at the tail.
This gradient serves as a guidance to optimize the latent denoising~\cite{yu2023freedom} as follows:
\begin{equation}
	\zkm \leftarrow \zkm - \gamma_t\cdot \nabla_{\zk} \sum_{t=1}^{T}\| \mathcal{K^\prime}_\text{ltm} - \mathcal{K^\prime}^{\state+t}_{t} \|_2^2~,
	\label{eq:loss_2}
\end{equation}
where $\gamma_t$ denotes a time-dependent strength of the guidance. By integrating this term, the self-recurrent guidance aligns the generated latents more closely with the subject features of the previously generated frames,  thereby enhancing long-range subject consistency in the synthesized video.

\section{Experiments}
\subsection{Experimental Settings}

\paragraph{Benchmark.} 
We empirically verify the merit of our Ouroboros-Diffusion for both single-scene and multi-scene long video generation on the VBench~\cite{huang2024vbench} benchmark.
We sample 93 common prompts from VBench as the testing set for single-scene video generation. 
All methods are required to generate 128 video frames for each prompt. 
To explore multi-prompt scenarios, we extended the single prompts into multiple prompts using a GPT-4o~\cite{openai2023gpt4}, resulting in 78 groups of multi-scene prompts.
Each group contains 2 to 3 prompts with consistent subject phrasing.
For each multi-prompt group, we generate $256$ video frames for performance comparison.

\begin{figure*}[t!]
	\centering
	\includegraphics[width=0.95\linewidth]{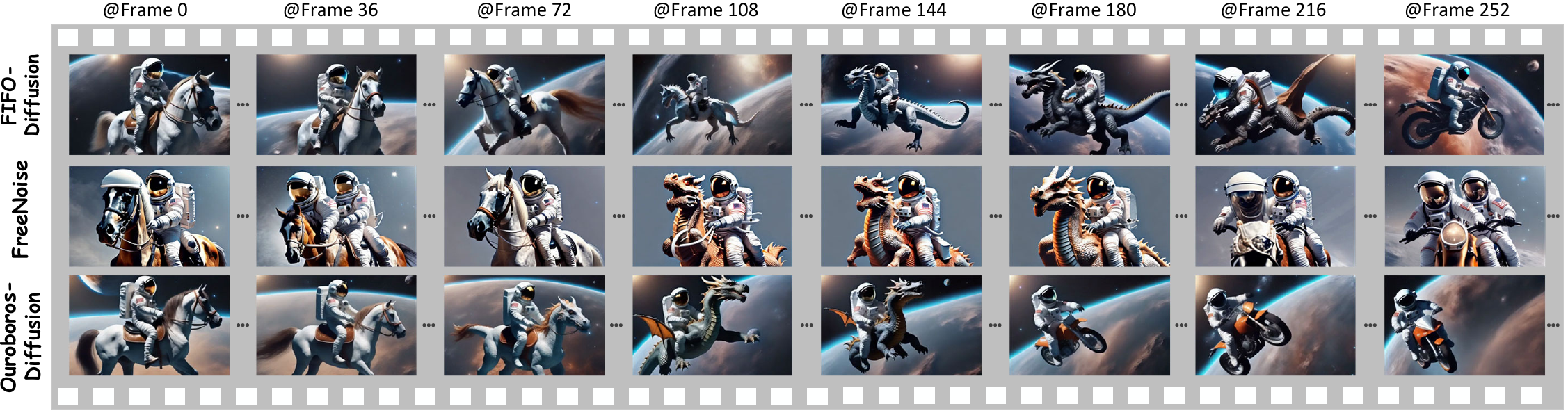}
	\caption{Visual examples of multi-scene long video generation by different approaches. The multi-scene prompts are: 1). an astronaut is riding a horse in space; 2). an astronaut is riding a dragon in space; 3). an astronaut is riding a motorcycle in space.}
	\label{fig:multi_case}
\end{figure*}

\begin{table*}[t]
	\centering
	\setlength{\tabcolsep}{0.8cm}{
		\resizebox{\linewidth}{!}{
			\begin{tabular}{lccccccc}
				\toprule
				\textbf{Approach} & 
				\makecell[c]{\textbf{Subject}\\ \textbf{Consistency}}$\uparrow$ & 
				\makecell[c]{\textbf{Background}\\ \textbf{Consistency}}$\uparrow$ & 
				\makecell[c]{\textbf{Motion}\\ \textbf{Smoothness}}$\uparrow$ & 
				\makecell[c]{\textbf{Temporal}\\ \textbf{Flickering}}$\uparrow$ & 
				\makecell[c]{\textbf{Aesthetic}\\ \textbf{Quality}}$\uparrow$ \\
				\midrule
				\multicolumn{1}{l|}{FIFO-Diffusion~\cite{kim2024fifo}} & 93.96 & 96.17 & 96.36 & 
				93.59 & 60.12 \\
				\multicolumn{1}{l|}{FreeNoise~\cite{qiu2023freenoise}} & 95.07 & 96.52 & 96.57 & 95.06 & \textbf{61.26} \\ \midrule
				\multicolumn{1}{l|}{Ouroboros-Diffusion} &  \textbf{95.73} & \textbf{96.82} &  \textbf{97.77} &  \textbf{95.82} & 61.17 \\
				\bottomrule
			\end{tabular}
	}}
	\caption{Multi-scene video generation performances on VBench. For each video, 256 frames are synthesized for evaluation.}
	\label{tab:me_exp}
\end{table*}

\paragraph{Implementation Details.} 
We implement our Ouroboros-Diffusion on the text-to-video model VideoCrafter2~\cite{chen2024videocrafter2}. The total number of time steps \(T\) in the DDIM sampler is set to 64, matching the queue length. The threshold for the low-pass filter in coherent tail latent sampling is set to 0.25. SACFA is applied only in the down-blocks and mid-block (with down-sampling factors of 2 and 4) of the spatial-temporal UNet empirically.
The last $16$ frames in the queue are involved in SACFA calculation.
The self-recurrent guidance derived from the first $16$ frames at the queue head applies to the last $16$ frames at the tail. 
The parameter \(\lambda\) for updating the subject feature bank is set to 0.98. 

\paragraph{Evaluation Metrics.}
We choose five evaluation metrics from VBench~\cite{huang2024vbench} for performance comparison: Subject Consistency, Background Consistency, Motion Smoothness, Temporal Flickering, and Aesthetic Quality.
Subject Consistency assesses the uniformity and coherence of the primary subject across frames using DINO~\cite{zhang2022dino} features.
Background Consistency is measured by the CLIP~\cite{radford2021learning} feature similarity.
Temporal Flickering evaluates the frame-wise consistency and Motion Smoothness assesses the fluidity and jittering of motion.
Finally, Aesthetic Quality indicates the quality of overall visual appearance including composition and color harmony.

\subsection{Comparisons with State-of-the-Art Methods}

We compare our proposal with four state-of-the-art long video diffusion models, \ie, StreamingT2V~\cite{henschel2024streamingt2v}, StreamingT2V-VideoTetris~\cite{tian2024videotetris}, FIFO-Diffusion~\cite{kim2024fifo} and FreeNoise~\cite{qiu2023freenoise} on long video generation.

\begin{table*}[t]
	\centering
        \fontsize{10pt}{12pt}\selectfont 
	\setlength{\tabcolsep}{0.4cm}{
		\resizebox{\linewidth}{!}{
			\begin{tabular}{c|ccc|ccccc}
				\toprule
				{\textbf{Model}} &
				\makecell[c]{\textbf{Coherent Tail}\\ \textbf{Latent Sampling}} & 
				\makecell[c]{\textbf{SACFA}} &
				\makecell[c]{\textbf{Self-Recurrent}\\ \textbf{Guidance}} & 
				\makecell[c]{\textbf{Subject}\\ \textbf{Consistency}}$\boldsymbol\uparrow$ & 
				\makecell[c]{\textbf{Background}\\ \textbf{Consistency}}$\boldsymbol\uparrow$ & 
				\makecell[c]{\textbf{Motion}\\ \textbf{Smoothness}}$\boldsymbol\uparrow$ & 
				\makecell[c]{\textbf{Temporal}\\ \textbf{Flickering}}$\boldsymbol\uparrow$ &
				\makecell[c]{\textbf{Aesthetic}\\ \textbf{Quality}}$\boldsymbol\uparrow$ \\
				\midrule
				A & - & - & \multicolumn{1}{c|}- & 94.04 & 96.08     & 95.88 & 93.38 & 59.06\\
				B & \checkmark & - & \multicolumn{1}{c|}- & 95.56     & 96.66     & 97.61 & 95.86 & 59.61 \\
				C & \checkmark & \checkmark & \multicolumn{1}{c|}- & 95.71   & {96.73}   & {97.70} & 96.00 & 59.67\\
				D & \checkmark & \checkmark & \multicolumn{1}{c|}\checkmark & \textbf{96.06}   & \textbf{96.90}      & \textbf{97.73}  & \textbf{96.12} & \textbf{59.89}\\
				\bottomrule
			\end{tabular}
	}}
	\caption{Performance contribution of each component (i.e., Coherent Tail Latent Sampling, SACFA and Self-Recurrent Guidance) in Ouroboros-Diffusion on single-scene video generation. For each video, 128 frames are synthesized for evaluation.}
	\label{tab:ablation}
\end{table*}

\paragraph{Single-Scene Video Generation.}
Table \ref{tab:sig_text_exp} summarizes the performance comparison of single-scene long video generation on VBench.
Overall, Ouroboros-Diffusion consistently outperforms other baselines across various metrics. Notably, Ouroboros-Diffusion achieves a Temporal Flickering score of $96.12\%$, surpassing the tuning-free approaches FIFO-Diffusion and FreeNoise by $2.74\%$ and $2.50\%$, respectively.
The highest frame consistency, as indicated by the Temporal Flickering metric, demonstrates the effectiveness of our coherent tail latent sampling, which enforces similarity in image layout between adjacent frames to enhance structural consistency.
Additionally, the best performances of Subject Consistency ($96.06\%$) and Background Consistency ($96.90\%$) further show that Ouroboros-Diffusion benefits from subject-level guidance, resulting in natural coherence throughout long video generation.
It is important to note that our approach does not compromise video motion strength (\eg, causing static video generation) to improve temporal consistency.
To validate this, we calculate the dynamic degree of the pre-trained base model VideoCrafter2~\cite{chen2024videocrafter2}, which achieves a score of 42.01. Ouroboros-Diffusion attains a higher dynamic degree of 44.12, confirming that our method maintains motion variability while further enhancing content consistency in video synthesis.

Figure \ref{fig:single_case} showcases a single-scene long video generation results across different approaches. Compared to other baselines, Ouroboros-Diffusion consistently produces videos with more seamless transitions and superior visual consistency. For example, StreamingT2V and FreeNoise often generate unreasonable or inconsistent content (\eg, the change of red collar in FreeNoise).
Although the FIFO-Diffusion denoising strategy maintains some content coherence (\eg, the appearance of the cat), the enqueued independent Gaussian noise still results in background variations (\eg, changes in the building behind the pool). In contrast, the video generated by our Ouroboros-Diffusion preserves both subject and background consistency effectively, demonstrating the advantage of using information guidance within the queue to enhance visual alignment in diffusion.

\begin{table}
	\centering
	\fontsize{10pt}{12pt}\selectfont
	\setlength{\tabcolsep}{0.1cm}{
		\resizebox{\linewidth}{!}{
			\begin{tabular}{lcc}
				\toprule
				\textbf{Model} & 
				\makecell[c]{\textbf{Motion} \textbf{Smoothness}}$\boldsymbol\uparrow$ & 
				\makecell[c]{\textbf{Temporal} \textbf{Flickering}}$\boldsymbol\uparrow$ \\
				\midrule
				\multicolumn{1}{l|}{Gaussian Noise}    & 95.88 & 93.38 \\ 
				\multicolumn{1}{l|}{Head Frame}  & 97.51 & 95.87 \\ \midrule
				\multicolumn{1}{l|}{Second-to-Last Frame (Ours)}     & \textbf{97.73} & \textbf{96.12} \\
				\bottomrule
			\end{tabular}
	}}
	\caption{Evaluation on the coherent tail latent sampling.}
	\label{tab:tail_ablation}
\end{table}

\paragraph{Multi-Scene Video Generation.}
Next, we compare our Ouroboros-Diffusion on the task of multi-scene video generation. Table \ref{tab:me_exp} details the performance across different baselines on VBench. Ouroboros-Diffusion outperforms all baselines in Subject/Background Consistency, Motion Smoothness, and Temporal Flickering.
Specifically, our approach demonstrates substantial performance boosts ($0.66\%$$\sim$$1.77\%$) in Subject Consistency.
Note that FreeNoise exploits a noise scheduler for multi-scene video generation, but it emphasizes on prompt adjustment in different denoising steps for motion injection.
Our Ouroboros-Diffusion differs fundamentally since ours not only integrates subject visual tokens into cross-frame attention for local alignment but also leverages these moving-averaged tokens to recurrently optimize video latents, ensuring global coherence.
Our Aesthetic Quality is slightly lower (by $0.09\%$) than that of FreeNoise.
We speculate that this may be due to a discrepancy between the parallel and diagonal denoising approaches (\ie, consistent noise versus inconsistent noise).
This issue could be addressed through model training, and it shows a direction for our future work.
Figure \ref{fig:multi_case} further illustrates the multi-scene long video generation results of three different approaches. As shown, Ouroboros-Diffusion successfully generates smoother transitions (\eg, scene changes with the astronaut maintaining the same motion direction) and more consistent visual content (\eg, a single, identical astronaut rather than two).

\subsection{Ablation Study on Ouroboros-Diffusion}
In this section, we conduct ablation studies to evaluate the impact of each design component in Ouroboros-Diffusion for long video generation. All experiments follow previous single-scene video generation settings for comparison.

\paragraph{Overall Framework.}
We first investigate how each component of the overall framework impacts the quality of video generation.
Table \ref{tab:ablation} summarizes the performance results for single-scene long video generation.
When integrating coherent tail latent sampling into the base model (\textbf{A}), a significant performance boost ($1.52\%$) is attained by model \textbf{B} in Subject Consistency.
This highlights a weakness of the base model (\ie, FIFO-Diffusion), where structural information may be overlooked due to the enqueueing of independent Gaussian noise at the queue tail.
By enhancing subject token alignment through subject-aware cross-frame attention, model \textbf{C} outperforms model \textbf{B} across all metrics.
Finally, model \textbf{D}, (\ie, our Ouroboros-Diffusion), achieves the best performance by recurrently propagating the subject information of frames from the queue head to the tail frames for latent optimization during video denoising.

\paragraph{Coherent Tail Latent Sampling.}
Next, we present the performance of different variants explored in the design of coherent tail latent sampling. Table \ref{tab:tail_ablation} details the results of two additional runs:1). enqueueing independent Gaussian noise at the tail, and 2) replacing the second-to-last frame with the head frame for latent sampling.

\begin{table}[h]
	\centering
	\fontsize{10pt}{12pt}\selectfont
	\setlength{\tabcolsep}{0.2cm}{
		\resizebox{\linewidth}{!}{
			\begin{tabular}{lcc}
				\toprule
				\textbf{Model} & 
				\makecell[c]{\textbf{Subject} \textbf{Consistency}}$\boldsymbol\uparrow$ & 
				\makecell[c]{\textbf{Motion} \textbf{Smoothness}}$\boldsymbol\uparrow$ \\
				\midrule
				w/o Guidance     & 94.04 & 95.88 \\
				Guidance with $\lambda$=1 & 95.87 & 97.71 \\
				Guidance with $\lambda$=0 & 96.00 & 97.71 \\ \midrule
				Moving-Average (Ours) & \textbf{96.06} & \textbf{97.73} \\
				\bottomrule
			\end{tabular}
	}}
	\caption{Evaluation on the self-recurrent guidance.}
	\label{tab:guide_ablation}
\end{table}

As expected, independent Gaussian noise yields the lowest Motion Smoothness. Using the head frame’s low-frequency component for latent sampling improves it from  $95.88\%$ to $97.51\%$. However, the appearance gap between queue head and tail latents limits structural guidance. By adjusting enqueued latents with tail information, Ouroboros-Diffusion further enhances motion quality.

\paragraph{Self-Recurrent Guidance.}
We have also analyzed the updating mechanism of the subject feature bank when devising the self-recurrent guidance. 
As shown in Table \ref{tab:guide_ablation}, whether using historic frame guidance ($\lambda$=1) or the current frame guidance ($\lambda$=0) at the queue head, there are notable performance improvements in Subject Consistency and Motion Smoothness.
These results highlight the advantage of guiding the denoising process using subject information through gradient-based adjustments.
To achieve better subject consistency in the synthesized video, we implement a moving-average strategy for feature bank updating, which combines information from both historic and current subject tokens.
The $\lambda$ is empirically set as $0.98$ in our framework.

\section{Conclusions}
This paper addresses consistent content generation in tuning-free long video diffusion. We introduce Ouroboros-Diffusion, a framework based on the first-in-first-out sampling strategy, which maintains a queue for frame-wise denoising. Our approach examines temporal consistency from two perspectives: structural and subject levels.
To materialize this idea, we inject structural information into newly enqueued Gaussian noise by leveraging the low-frequency component of latents near the queue tail, thereby enhancing the consistency of the overall structural layout. At the subject level, we emphasize short-range consistency through visual token alignment in cross-frame attention, while long-range consistency is achieved via latent guidance using subject-aware gradient adjustment.
Experiments on the VBench benchmark validate the effectiveness of Ouroboros-Diffusion, demonstrating improvements in both visual quality and motion smoothness.

\bibliography{aaai25}

\end{document}